\pdfoutput=1
\documentclass[11pt]{article}

\usepackage[preprint]{acl}
\usepackage{array} 
\usepackage{times}
\usepackage{latexsym}
\usepackage{url}
\usepackage{hyperref}
\usepackage{tabularx}
\usepackage[table,xcdraw,dvipsnames]{xcolor}
\usepackage{multirow}
\usepackage{array}
\usepackage{enumitem}
\usepackage{float}
\usepackage{adjustbox}
\usepackage{tcolorbox}
\usepackage{booktabs} % For better-looking tables
\usepackage{amsfonts}
\usepackage[T1]{fontenc}
\usepackage[utf8]{inputenc}
\usepackage{microtype}
\usepackage{inconsolata}
\usepackage{graphicx}
\usepackage{xspace}
\usepackage{booktabs}
\usepackage{makecell}
\usepackage{multirow}
\usepackage{caption}

\setlist{leftmargin=*,nosep}

\newcommand{\nm}[0]{{\tt NyayaMind}\xspace}
\newcommand{\nmbf}[0]{{\tt \textbf{NyayaMind}}\xspace}
\newcommand{\nmdb}[0]{{\tt NyayaMind-DB}\xspace}
\newcommand{\nmdbbf}[0]{{\tt \textbf{NyayaMind-DB}}\xspace}
\newcommand{\lseg}[0]{{\tt LegalSeg}\xspace}
\newcommand{\lsegbf}[0]{{\tt \textbf{LegalSeg}}\xspace}

\title{NyayaMind: A Framework for Transparent Legal Reasoning and Judgment Prediction in the Indian Legal System}

\author{Parjanya Aditya Shukla$^{1*}$ \quad {Shubham Kumar Nigam}$^{1,5* \dagger}$ \quad Debtanu Datta$^{2}$ \\ \textbf{Balaramamahanthi Deepak Patnaik}$^{1}$ \quad
\textbf{Noel Shallum}$^{3}$ \quad \textbf{Pradeep Reddy Vanga}$^{4}$ \\ \textbf{Saptarshi Ghosh}$^{2}$ \quad \textbf{Arnab Bhattacharya}$^{1}$\\
$^{1}$ IIT Kanpur, India \quad $^{2}$ IIT Kharagpur, India \quad $^{3}$ Symbiosis Law School Pune, India \\
$^{4}$ Dattam Labs, India \quad
$^{5}$ University of Birmingham, Dubai, United Arab Emirates \\
\texttt{\{padityashukla26, shubhamkumarnigam, debtanudatta04, bdeepakpatnaik2002} \\\texttt{noelshallum, pradeepreddy.vanga\}@gmail.com} \\\texttt{saptarshi@cse.iitkgp.ac.in} \quad \texttt{arnabb@iitk.ac.in} 
}%author
\begin{document}
\maketitle

\renewcommand{\thefootnote}{$*$}
\footnotetext{These authors contributed equally to this work}
\renewcommand{\thefootnote}{$\dagger$}
\footnotetext{Corresponding author}
\renewcommand{\thefootnote}{\arabic{footnote}}

\begin{abstract}
Court Judgment Prediction and Explanation (CJPE) aims to predict a judicial decision and provide a legally grounded explanation for a given case based on the facts, legal issues, arguments, cited statutes, and relevant precedents. For such systems to be practically useful in judicial or legal research settings, they must not only achieve high predictive performance but also generate transparent and structured legal reasoning that aligns with established judicial practices. 
In this work, we present \nmbf, an open-source framework designed to enable transparent and scalable legal reasoning for the Indian judiciary. The proposed framework integrates retrieval, reasoning, and verification mechanisms to emulate the structured decision-making process typically followed in courts. Specifically, \nm consists of two main components: a \textit{Retrieval Module} and a \textit{Prediction Module}. The \textit{Retrieval Module} employs a RAG pipeline to identify legally relevant statutes and precedent cases from large-scale legal corpora, while the \textit{Prediction Module} utilizes reasoning-oriented LLMs fine-tuned for the Indian legal domain to generate structured outputs including issues, arguments, rationale, and the final decision. 
Our extensive results and expert evaluation demonstrate that \emph{\nm significantly improves the quality of explanation and evidence alignment compared to existing CJPE approaches}, providing a promising step toward trustworthy AI-assisted legal decision support systems.
\end{abstract}

% Court Judgment Prediction and Explanation (CJPE) aims to predict a decision (with explanation) for a given legal case based on the facts, legal issues, arguments, cited statutes, and semantically retrieved prior cases. For a CJPE system to be used in courts, it must be reliable in both prediction accuracy and legal explanation. In this study, we propose \nmbf, a CJPE system backed by reasoning-based Large Language Models (LLMs) that provides a decision along with supporting explanation, following a legal reasoning process. Our \nm consists of two distinct components -- the \textit{Prediction} and the \textit{Retrieval} modules. The \textit{Prediction} module employs fine-tuned reasoning-based LLMs tailored for judgment prediction in the Indian legal domain, whereas the \textit{Retrieval} Module uses Retrieval-Augmented Generation (RAG) to retrieve pertinent statutes and precedents. Our findings, validated through domain experts, showcase that \dd{@Noel: Pls add the main findings here.}

\section{Introduction}
\label{sec:intro}

%%%%%%%%%%%%%%%%%%%%%%%%%%%%%%%%%%%%%%%%%%%%%%%%%%%%%%%%%%%%
\begin{figure*}[t]
    \centering 
    \includegraphics[width=0.95\linewidth]{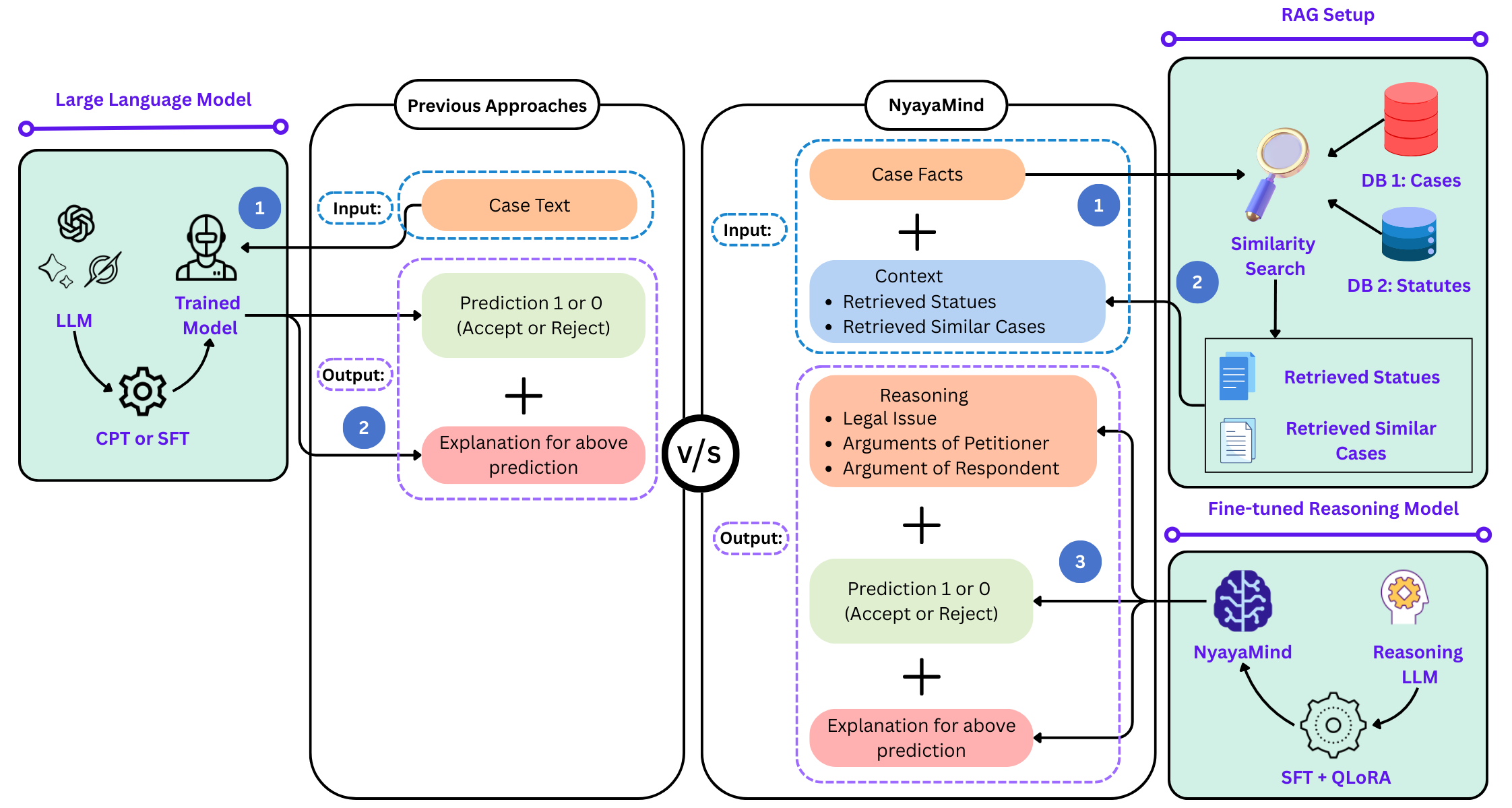} 
    \caption{Comparison of our \nmbf System with existing prior approaches.}
    \label{fig:NyayaMind_Task_Diagram}
\end{figure*}
%%%%%%%%%%%%%%%%%%%%%%%%%%%%%%%%%%%%%%%%%%%%%%%%%%%%%%%%%%%%

Judicial decision-making is an inherently structured process in which courts analyze statutory provisions, examine precedents, evaluate competing arguments from different parties, and ultimately deliver a reasoned decision grounded in legal principles. In practice, faithfully following this deliberative process -- identifying legal issues, weighing arguments, grounding conclusions in statutes and precedents, and articulating coherent reasoning -- requires significant time and expertise. The challenge becomes even more pronounced in large judicial systems such as India, where over 50 million cases remain pending as of March 2026\footnote{\url{https://ecourts.gov.in/ecourts_home/}}. The scale of this backlog places immense pressure on courts, often limiting the time available for detailed legal analysis. As a result, there is growing interest in developing AI systems that can assist legal professionals in structuring and analyzing legal information while preserving the transparency and rigor required in judicial decision-making.

Recent advances in LLMs have shown strong potential in complex reasoning and domain-specific language understanding, leading to increasing exploration of LLMs for legal judgment prediction and explanation. Prior work has introduced domain-specific datasets and models for the legal domain, including \textit{PredEx}~\citep{nigam-etal-2024-legal} and \textit{INLegalLlama} built on the \textit{NyayaAnumana} corpus~\citep{nigam-etal-2025-nyayaanumana}. These studies demonstrate that LLMs can capture legal semantics and generate explanations for predicted judgments. Other approaches have incorporated retrieval-based techniques, such as RAG, to integrate statutes and precedents into model predictions. While these methods improve contextual grounding, most existing approaches primarily optimize for outcome prediction or surface-level explanation generation rather than explicitly modeling the structured reasoning process followed by courts.

A fundamental limitation of current Legal AI systems is their lack of transparent and verifiable reasoning. LLMs often generate explanations that appear plausible but may not be faithfully grounded in the retrieved legal sources, leading to hallucinated rationales or incomplete legal reasoning. Moreover, existing approaches rarely model the full deliberative structure of judicial reasoning, such as identifying legal issues, articulating petitioner and respondent arguments, evaluating statutory provisions, and deriving conclusions through logical justification. As a result, many current systems behave as black-box predictors rather than transparent reasoning assistants capable of supporting legal analysis in practice. Figure~\ref{fig:NyayaMind_Task_Diagram} illustrates the key differences between our proposed framework and prior approaches.

To address these limitations, we introduce \nmbf, an open-source framework designed to enable transparent and scalable legal reasoning for the Indian judiciary. The proposed system integrates retrieval, reasoning, and verification components to emulate the structured reasoning process typically followed in judicial decision-making. \nm consists of two main modules: a \textit{Retrieval Module} and a \textit{Prediction Module}. The \textit{Retrieval Module} employs a RAG pipeline to identify relevant statutes and precedent cases based on the factual description of a legal dispute. The \textit{Prediction Module} utilizes reasoning-oriented LLMs fine-tuned for the Indian legal domain to generate structured legal outputs, including legal issues, petitioner arguments, respondent arguments, statutory grounding, and the final decision. During fine-tuning, the model is explicitly supervised to follow this deliberative reasoning schema, guiding it to emulate the argumentative structure commonly observed in judicial opinions.

Our key contributions in this work are as follows:

\begin{itemize}
    \item \textit{A Transparent Legal Reasoning Framework:} We introduce \nmbf, an open-source framework that integrates retrieval, structured reasoning, and verification mechanisms to support transparent legal decision analysis in the judicial context.

    \item \textit{Structured Reasoning with LLMs for CJPE:} To the best of our knowledge, this is one of the first works in the Indian legal domain to fine-tune reasoning-based LLMs to explicitly follow a structured judicial reasoning paradigm for the CJPE task.

    \item \textit{Introducing a Large-scale Legal Database:} We introduce \nmdbbf, a large-scale legal database for the Indian legal domain containing over 16M legal documents that can be leveraged to retrieve relevant statutes and precedents for a RAG-based retrieval module.

    \item \textit{Comprehensive Evaluation of Retrieval Pipelines:} We systematically evaluate multiple RAG-based retrieval pipelines for statutes and precedents using different vector DBs to analyze their effectiveness in supporting legal reasoning.

    \item \textit{Comparative Analysis of Reasoning LLMs:} We fine-tune and evaluate four reasoning-oriented LLMs ranging from 4B to 27B parameters, providing insights into their capability to generate structured legal reasoning in the legal domain.

    \item \textit{Detailed Ablation with Quantization:} We conduct detailed experiments using parameter-efficient LoRA fine-tuning under multiple quantization settings to analyze the tradeoffs between computational efficiency and reasoning quality.

    \item \textit{Expert Evaluation:} We conduct expert-based evaluation with legal practitioners to assess the coherence, factual grounding, and legal plausibility of the generated reasoning.

\end{itemize}
To ensure reproducibility and encourage further research, the dataset and model code will be made publicly available soon.
% Our codebase is available via this anonymous repository\footnote{\url{https://anonymous.4open.science/r/NyayaMind/}}.

\section{Related Work}

Legal Judgment Prediction (LJP) has long been studied as an important task in Legal AI, aiming to automatically predict court decisions based on case facts, legal arguments, and supporting evidence. Early work relied on case-based reasoning and symbolic legal models to predict outcomes and provide explanations \citep{10.1145/1047788.1047838}. With the rise of neural methods, researchers began incorporating legal knowledge into deep learning architectures to improve predictive performance and capture complex legal dependencies \citep{gan2021judgment, huang2021dependency}. More recent work has explored combining domain-specific models with large language models (LLMs) to utilize precedents and contextual legal information effectively \citep{wu-etal-2023-precedent}. 

Another key direction focuses on transparency and explainability, which are critical for deploying AI systems in high-stakes domains such as law. Legal scholars emphasize that AI systems must provide interpretable reasoning to ensure accountability and compliance with legal principles \citep{papadouli2022transparency, esposito2022transparency}. Several studies have proposed approaches to enhance explainability by incorporating structured legal information such as entities, citations, or representative prototypes into prediction models \citep{benedetto2025boosting, luo-etal-2023-prototype}. Policy discussions around emerging regulations, including the EU AI Act, further highlight the importance of transparency and traceability in AI-assisted legal decision-making \citep{makauskaite2025transparency}.

Recent advances in LLMs have significantly expanded the capabilities of legal reasoning systems. Frameworks such as LegalReasoner integrate knowledge retrieval and reasoning steps for improved judgment prediction \citep{10750819}, while discriminative reasoning frameworks help LLMs distinguish between similar legal charges and improve decision accuracy \citep{deng-etal-2024-enabling}. Other work explores integrating logical reasoning and semantic knowledge to enhance the reliability of LLM-generated legal responses \citep{yao-etal-2025-elevating}. Reinforcement learning and iterative reasoning approaches have also been proposed to strengthen legal reasoning abilities in LLMs, as demonstrated in UniLaw-R1 \citep{cai2025unilaw}. Additionally, emerging agent-based legal AI systems leverage dynamic reasoning and search mechanisms to solve complex legal tasks, such as LawThinker and LRAS \citep{yang2026lawthinker, zhou2026lras}. Benchmarks like LEXAM further highlight the need for robust evaluation of legal reasoning capabilities in AI systems \citep{fan2025lexam}. 

In the Indian legal domain, several datasets and models have been introduced to support research in judgment prediction and explanation. PredEx provides expert-annotated data for prediction and explanation tasks \citep{nigam-etal-2024-legal}, while NyayaAnumana and INLegalLlama offer large-scale corpora and domain-specific language models tailored to Indian legal texts \citep{nigam-etal-2025-nyayaanumana}. Furthermore, prior work has studied judgment prediction in realistic legal scenarios where models rely only on information available before the final verdict \citep{nigam-etal-2024-rethinking}. Despite these advances, many existing approaches focus either on prediction accuracy or explanation generation independently. 
% In contrast, our proposed \nm framework integrates retrieval, structured reasoning, and verification mechanisms within a unified pipeline to produce transparent and evidence-grounded legal reasoning for the Indian judiciary.

\section{Task Description}
\label{sec:task-desc}

The overall pipeline of \nm\ consists of four sequential sub-tasks: \textit{Retrieval}, \textit{Reasoning},  \textit{Prediction} and \textit{Explanation}. 

\subsection{Retrieval Task}
The Retrieval task is formulated as a binary classification problem. Given the case facts \(F\), a list of legal statutes \(L_s\), and a list of prior cases \(L_c\), the objective is to determine which documents are relevant to the current case. Formally, for each statute \( s_i \) and case  \( c_i \) such that \( s_i \in L_s \) and \( c_i \in L_c \) a label of \(1\) (accept) or \(0\) (reject) is generated. A label of \(1\) (accept) indicates that the document should be included in the set of retrieved documents \(L_d\), whereas a label of \(0\) (reject) indicates that the document should be excluded. The classification decision is based on the semantic similarity between the case facts \(F\) and the content of the candidate legal document. 

\subsection{Reasoning Task} Given the case facts \(F\) and the retrieved legal documents (statutes and precedents) \(L_d\), the system performs structured legal reasoning. Specifically, the model generates three key components: the \textit{legal issue} \(I\), the \textit{arguments presented by the petitioner} \(A_p\), and the \textit{arguments presented by the respondents} \(A_r\).
These elements together form a structured legal reasoning \(R\), where \(I, A_p, A_r \in R\). This intermediate reasoning step is essential for improving the transparency, interpretability, and reliability of the \nm\ system, as it explicitly showcases the argumentative structure underlying legal decisions.

\subsection{Prediction Task}
The Prediction task determines the outcome of the legal appeal. Given the case facts \(F\) and the structured reasoning representation \(R\), the model predicts a binary decision label \(y \in \{0,1\}\). A label \(0\) indicates that the appeal is rejected, whereas a label \(1\) indicates that the appeal is accepted.

This prediction stage leverages both the factual context and the generated reasoning structure, allowing the system to produce decisions that are informed by explicit legal analysis rather than purely statistical correlations.

\subsection{Explanation Task}
The final stage of the pipeline generates a comprehensive legal explanation \(E\) supporting the predicted outcome \(y\). The explanation integrates information from the case facts \(F\), the structured reasoning \(R\), and the retrieved legal documents \(L_d\). In particular, the generated explanation highlights the legal issues involved, summarizes the arguments presented by both parties, references relevant statutes and precedents, and articulates the rationale behind the final decision.

By jointly modeling retrieval, reasoning, prediction, and explanation, \nm\ aims to provide interpretable and legally grounded decision support. This structured formulation allows the system to produce transparent outputs that more closely resemble the reasoning process used by human judges in the Indian legal system.

\section{Dataset}
\label{sec:dataset}
Our study leverages two primary datasets for developing the \nm system: \nmdbbf and \lsegbf~\citep{nigam-etal-2025-legalseg}. The former is used to support large-scale retrieval of statutes and precedents, while the latter is used for fine-tuning reasoning-oriented LLMs for structured legal reasoning and judgment prediction.

\subsection{The \nmdbbf Dataset}
We construct \nmdbbf as a large-scale legal corpus tailored for the Indian judicial domain. The dataset comprises over 16 million legal documents, including approximately 60K judgments from the Supreme Court of India and nearly 16M judgments from multiple High Courts. In addition, it includes around 1K Central Acts and approximately 7K State Acts. The judicial documents are collected from the \textit{E-Courts} platform\footnote{\url{https://judgments.ecourts.gov.in/}}, while statutory documents are sourced from \textit{IndiaCode}\footnote{\url{https://indiacode.nic.in/}}.

The \nmdb serves as the backbone of our Retrieval module, enabling large-scale retrieval of legally relevant statutes and precedents using semantic similarity between case facts and candidate documents. Importantly, the dataset contains not only final judgments but also intermediate court orders associated with each case, thereby providing a richer representation of the judicial process. This diversity and scale allow the system to capture nuanced legal context and improve the grounding of generated reasoning.

\subsection{Dataset used for fine-tuning} 
For training the reasoning component of \nm, we utilize the \lseg dataset~\citep{nigam-etal-2025-legalseg}, which focuses on semantic segmentation of legal judgments through rhetorical role classification in the Indian legal context. The dataset consists of 7,120 cases, split into 4,984 training samples, 712 validation samples, and 1,424 test samples. Each case is annotated with structured components such as \textit{Facts}, \textit{Issues}, \textit{Reasoning}, \textit{Arguments of Petitioner}, \textit{Arguments of Respondent}, and \textit{Decision}.

To align with our task formulation, we transform the dataset into a structured input-output format suitable for reasoning-based LLMs. Specifically, we employ an LLM to rewrite the annotated fields into a conversational and coherent format, while preserving their legal semantics. The dataset also includes additional contextual information such as cited statutes (\textit{section\_texts}) and cited precedent cases (\textit{cited\_case\_judgments}), which are incorporated into the model inputs to strengthen legal grounding.

During fine-tuning, the model is trained to generate structured legal reasoning by first identifying issues, then articulating arguments from both parties, and finally producing a decision along with a justified explanation. The prompt templates used for different models are detailed in Tables~\ref{tab:prompt-deepseek-r1}, \ref{tab:prompt-phi4-mini-reasoning}, \ref{tab:prompt-phi4-reasoning}, and \ref{tab:prompt-qwen3.5} in the Appendix. Additionally, a detailed breakdown of token distribution across dataset components, including input and target texts, is provided in Table~\ref{tab:dataset-tokens}. This structured supervision enables the model to learn a faithful approximation of judicial reasoning rather than relying on shallow pattern matching.

\section{Methodology}

\begin{figure*}[t]
    \centering 
    \includegraphics[width=0.95\linewidth]{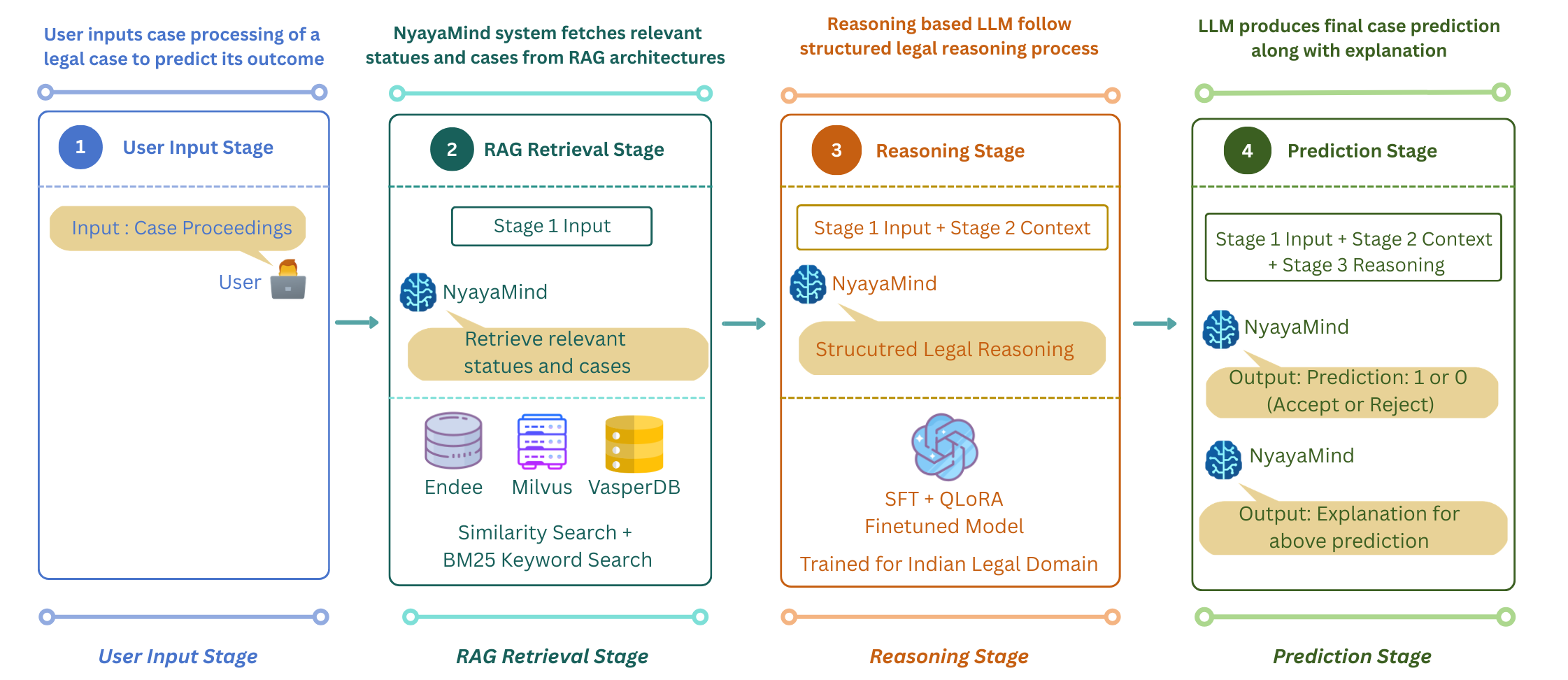} 
    \caption{Representation of different stages in the flow of our \nmbf System.}
    \label{fig:NyayaMind_Flow_Diagram}
\end{figure*}

The \nm framework is designed as a multi-stage pipeline that closely mirrors the structured reasoning process followed in judicial decision-making. As illustrated in Figure~\ref{fig:NyayaMind_Flow_Diagram}, the system consists of four sequential stages: \textit{User Input}, \textit{Retrieval}, \textit{Reasoning}, and \textit{Prediction \& Explanation}. This design ensures that the final decision is not generated in isolation, but is instead grounded in retrieved legal evidence and structured intermediate reasoning.

Given the case facts \(F\) as input, the first step involves retrieving relevant legal knowledge. To this end, we implement three distinct Retrieval-Augmented Generation (RAG) architectures: \textit{Milvus}, \textit{Endee}, and \textit{Vespa}. These systems differ in indexing strategies, similarity metrics, chunking mechanisms, and retrieval pipelines. Each RAG setup combines dense vector similarity search with lexical matching (BM25), enabling robust retrieval of statutes and precedent cases that are semantically and contextually aligned with the input case. This multi-system design allows us to systematically evaluate the impact of retrieval quality on downstream reasoning and prediction. For reproducibility, we provide detailed configurations of indexing, retrieval parameters, and integration pipelines in Appendix~\ref{sec:implementation}, with a comparative summary presented in Table~\ref{tab:rag-comparison-wide}.

The retrieved documents \(L_d\), along with the original case facts \(F\), are then provided as input to the reasoning component. To ensure that the generated outputs follow a principled legal structure, we fine-tune reasoning-oriented Large Language Models (LLMs) using Supervised Fine-Tuning (SFT) combined with parameter-efficient techniques such as LoRA~\citep{hu2021loralowrankadaptationlarge} and QLoRA~\citep{dettmers2023qloraefficientfinetuningquantized}. During training, the models are explicitly guided to follow a structured legal reasoning schema, which includes identifying legal issues, articulating arguments from the petitioner and respondent, grounding reasoning in statutory provisions, and deriving a final decision. This process is implemented through structured prompting and supervision within dedicated reasoning segments (``\textbf{think tokens}''), encouraging the model to perform step-by-step deliberation rather than direct answer generation.

Finally, the model produces the prediction \(y\) (accept/reject) along with a detailed explanation \(E\) that is grounded in both the retrieved legal evidence and the generated reasoning structure. By tightly coupling retrieval, reasoning, and prediction, \nm ensures that the generated outputs are not only accurate but also interpretable and verifiable. This integrated design enables \nm to function as a transparent legal reasoning system, bridging the gap between black-box prediction models and the structured analytical processes required in real-world judicial settings.

\section{Experimental Setup}
\label{sec: experiment}

This section describes the experimental setup used to adapt reasoning-based LLMs for the CJPE task in the Indian legal domain.

\subsection{Models}
We evaluate four recent reasoning-oriented or instruction-tuned LLMs of varying scales: 
(i) \textit{DeepSeek-R1-Distill-Qwen-14B}\footnote{\scriptsize \url{https://huggingface.co/deepseek-ai/DeepSeek-R1-Distill-Qwen-14B}}, 
(ii) \textit{Phi-4-mini-reasoning}\footnote{\scriptsize \url{https://huggingface.co/microsoft/Phi-4-mini-reasoning}}, 
(iii) \textit{Phi-4-reasoning}\footnote{\scriptsize \url{https://huggingface.co/microsoft/Phi-4-reasoning}}, and 
(iv) \textit{Qwen3.5-27B}\footnote{\scriptsize \url{https://huggingface.co/Qwen/Qwen3.5-27B}}. 
These models were selected to compare different model families and parameter scales in terms of their ability to perform structured legal reasoning, outcome prediction, and explanation generation.

\subsection{Training and Inference Setup}
Our training setup is designed to adapt reasoning LLMs to the CJPE task while maintaining computational efficiency and preserving their capacity for multi-step structured legal analysis.

\noindent {\textbf{Parameter-efficient Training}}:  
Instead of full-parameter fine-tuning, we adopt a parameter-efficient fine-tuning strategy based on \textit{LoRA}~\citep{hu2021loralowrankadaptationlarge} and \textit{QLoRA}~\citep{dettmers2023qloraefficientfinetuningquantized}. LoRA adapters are inserted into both attention projection layers and feed-forward projection layers, allowing the models to specialize to the legal domain with substantially reduced memory overhead. This setup enables efficient adaptation of large reasoning models while retaining their general reasoning capabilities.

\noindent {\textbf{Quantization Strategy}}:  
To study the trade-off between efficiency and reasoning quality, we experiment with multiple precision settings for \textit{DeepSeek-R1-Distill-Qwen-14B}, namely 4-bit, 8-bit, 16-bit, and 32-bit quantization. The 4-bit version uses the \textit{DeepSeek-R1-Distill-Qwen-14B-unsloth-bnb-4bit}\footnote{\scriptsize \url{https://huggingface.co/unsloth/DeepSeek-R1-Distill-Qwen-14B-unsloth-bnb-4bit}} variant provided through the \textit{Unsloth}\footnote{\scriptsize \url{https://unsloth.ai/}} framework. In contrast, \textit{Phi-4-mini-reasoning}, \textit{Phi-4-reasoning}, and \textit{Qwen3.5-27B} are fine-tuned in 16-bit precision. This design enables a systematic comparison of memory usage, training feasibility, and downstream reasoning fidelity across quantization levels. For \textit{Qwen3.5-27B}, we further use a longer context window of 32K tokens to better accommodate lengthy legal inputs.

\noindent {\textbf{Hyperparameter Configuration}}:  
Training is implemented using the Unsloth framework with distributed training support for efficient optimization. We use gradient accumulation to handle long-context inputs under constrained GPU memory, and apply optimized attention implementations for selected models to improve throughput. The complete training and inference hyperparameters are reported in Table~\ref{tab:training-parameters} in the Appendix.

\noindent {\textbf{Prompt Design for Legal Reasoning}}:  
To guide the models toward structured judicial reasoning, we design task-specific prompts tailored to the CJPE setting. The prompts integrate multiple legally relevant fields from the training data, including simplified case facts, cited statutory provisions, cited prior judgments, legal issues, arguments from the petitioner and respondent, the final decision, and the associated judicial reasoning. Rather than directly predicting the outcome from raw facts, the model is trained to follow an explicit reasoning trajectory through dedicated intermediate ``think tokens,'' which correspond to key legal reasoning steps such as issue identification and argument analysis. The model is then trained to generate both the final decision and a supporting natural language explanation. Prompt templates for all model variants are provided in Tables~\ref{tab:prompt-deepseek-r1}, \ref{tab:prompt-phi4-mini-reasoning}, \ref{tab:prompt-phi4-reasoning}, and \ref{tab:prompt-qwen3.5}.

\noindent \textbf{Inference Stage}:  
At inference time, \nm takes as input the factual description of a legal case. This input is augmented with relevant statutes and precedent cases retrieved through our RAG-based retrieval module, which combines semantic and lexical retrieval strategies over Indian legal corpora. The retrieved legal context is then passed to the fine-tuned reasoning model, which generates structured intermediate reasoning followed by the final judicial outcome and its explanation. In this way, the system grounds its prediction in both retrieved legal evidence and an explicit reasoning process, improving transparency and legal plausibility. The complete inference pipeline is illustrated in Figure~\ref{fig:NyayaMind_Flow_Diagram}.

\section{Evaluation Metrics}
\label{sec:eval-metrics}

We evaluate the performance of \nm using both automatic metrics and expert-based human evaluation to comprehensively assess prediction accuracy, reasoning quality, and explanation fidelity.

\noindent {\textbf{Automatic Evaluation:}}  
We employ a combination of lexical and semantic similarity metrics to evaluate the generated outputs. For lexical overlap, we report \textit{ROUGE} scores (ROUGE-1, ROUGE-2, and ROUGE-L) \citep{lin-2004-rouge}, \textit{BLEU} \citep{papineni2002bleu}, and \textit{METEOR} \citep{banerjee-lavie-2005-meteor}. These metrics capture n-gram overlap, fluency, and word-order similarity between the generated outputs and reference texts. 

To assess semantic alignment, we use \textit{BERTScore} \citep{BERTScore}, which computes contextual similarity between generated and reference texts using pretrained language model embeddings. Additionally, we report \textit{BLANC} \citep{blanc}, a reference-free metric that evaluates how well the generated text supports understanding of the original input. Together, these metrics provide a balanced evaluation of both surface-level correctness and deeper semantic consistency of the generated explanations.

\noindent {\textbf{Expert Evaluation:}}  
To evaluate the legal validity and interpretability of the generated outputs, we conduct an expert evaluation with two legal professionals. The evaluation focuses on both intermediate reasoning components, \textit{legal issue}, \textit{arguments of petitioner}, and \textit{arguments of respondent}, and final outputs, \textit{prediction} and \textit{explanation}. Each component is rated on a 1--10 Likert scale based on criteria such as accuracy, relevance, coherence, and completeness. In addition to component-wise scores, an overall score is assigned to reflect the holistic quality of the system output. A score of 1 indicates poor or irrelevant output, whereas a score of 10 indicates high-quality reasoning and explanation, potentially comparable to or exceeding the reference text. This expert-driven evaluation complements automatic metrics by capturing domain-specific correctness and practical usefulness in the Indian legal context.

\section{Result and Analysis}
% Judgement prediction and explanation results
%%%%%%%%%%%%%%%%%%%%%%%%%%%%%%%%%%
\begin{table}[t]
\centering
\small
\setlength{\tabcolsep}{5pt}
\renewcommand{\arraystretch}{1.15}

\resizebox{\linewidth}{!}{
\begin{tabular}{lcccccccc}
\toprule

& \multicolumn{5}{c}{\textbf{Lexical Evaluation (\%)}} 
& \multicolumn{2}{c}{\textbf{Semantic Evaluation (\%)}} 
& \textbf{Expert} \\

\cmidrule(lr){2-6} \cmidrule(lr){7-8}

\textbf{Models}
& \textbf{R-1} 
& \textbf{R-2} 
& \textbf{R-L} 
& \textbf{BLEU} 
& \textbf{METEOR}
& \textbf{BERTScore}
& \textbf{BLANC}
& \textbf{Rating} \\

\midrule
\multicolumn{9}{c}{\textbf{Generated Legal Issue vs Ground Truth}} \\
\midrule

\texttt{DeepSeek-R1-Distill-Qwen-14B}
& 27.06 & 8.99 & 18.81 & 3.12 & 16.87 & 52.98 & 7.75 & 6.25 \\

\texttt{Qwen3.5-27B}
& \textbf{27.73} & \textbf{9.68} & \textbf{19.23} & \textbf{3.56} & \textbf{17.84} & \textbf{53.21} & \textbf{8.37} & \textbf{6.95} \\

\midrule
\multicolumn{9}{c}{\textbf{Generated Petitioner Arguments vs Ground Truth}} \\
\midrule

\texttt{DeepSeek-R1-Distill-Qwen-14B}
& 30.65 & 10.08 & 18.30 & 3.90 & \textbf{22.19} & 57.09 & \textbf{9.85} & 6.50 \\

\texttt{Qwen3.5-27B}
& \textbf{32.18} & \textbf{11.29} & \textbf{19.20} & \textbf{4.22} & 21.23 & \textbf{58.25} & 9.36 & \textbf{6.90} \\

\midrule
\multicolumn{9}{c}{\textbf{Generated Respondent Arguments vs Ground Truth}} \\
\midrule

\texttt{DeepSeek-R1-Distill-Qwen-14B}
& 24.59 & 6.95 & 15.80 & 2.07 & \textbf{16.44} & 53.95 & 5.91 & 6.20 \\

\texttt{Qwen3.5-27B}
& \textbf{25.91} & \textbf{8.08} & \textbf{16.77} & \textbf{2.38} & 15.93 & \textbf{55.20} & \textbf{7.14} & \textbf{6.90} \\

\midrule
\multicolumn{9}{c}{\textbf{Generated Prediction vs Ground Truth}} \\
\midrule

\texttt{DeepSeek-R1-Distill-Qwen-14B}
& 43.89 & 27.22 & 36.67 & 18.27 & 35.70 & 66.59 & \textbf{22.89} & 5.50 \\

\texttt{Qwen3.5-27B}
& \textbf{46.63} & \textbf{28.57} & \textbf{38.31} & \textbf{20.14} & \textbf{36.73} & \textbf{67.30} & 22.73 & \textbf{6.90} \\

\midrule
\multicolumn{9}{c}{\textbf{Generated Explanation vs Ground Truth}} \\
\midrule

\texttt{DeepSeek-R1-Distill-Qwen-14B}
& \textbf{32.80} & \textbf{12.63} & \textbf{18.64} & \textbf{4.39} & 18.05 & 60.70 & 8.25 & 6.12 \\

\texttt{Qwen3.5-27B}
& 29.95 & 11.89 & 17.04 & 4.35 & \textbf{19.83} & \textbf{61.90} & \textbf{9.77} & \textbf{7.00} \\

\bottomrule
\end{tabular}
}

\caption{Performance comparison of models on legal judgment explanation tasks. Lexical metrics include ROUGE, BLEU, and METEOR, while semantic similarity is measured using BERTScore and BLANC. Expert scores reflect human evaluation by legal professionals. Best scores are shown in bold.}
\label{tab:explanation-table}
\end{table}
%%%%%%%%%%%%%%%%%%%%%%%%%%%%%%%%%%%%%

%  IAA Scores
%%%%%%%%%%%%%%%%%%%%%%%%%%%%%%%%%%%%%%%%%%%%%%%
\begin{table}[t]
\centering
\small
\resizebox{\columnwidth}{!}{
\setlength{\tabcolsep}{4pt}
\renewcommand{\arraystretch}{1.12}

\begin{tabular}{p{0.46\columnwidth}ccc}
\toprule
\textbf{Generated Component} & \textbf{Expert 1} & \textbf{Expert 2} & \textbf{Average} \\
\midrule

\multicolumn{4}{c}{\textbf{\texttt{DeepSeek-R1-Distill-Qwen-14B} (32-bit)}} \\
\addlinespace
Legal Issue & 6.48 & 6.00 & 6.24 \\
Arguments of Petitioner & 6.44 & 6.75 & 6.60 \\
Arguments of Respondent & 6.12 & 6.83 & 6.48 \\
Prediction & 6.60 & 6.25 & 6.43 \\
Explanation & 5.88 & 5.89 & 5.88 \\

\midrule

\multicolumn{4}{c}{\textbf{\texttt{Qwen3.5-27B}}} \\
\addlinespace
Legal Issue & 6.40 & 7.80 & 7.10 \\
Arguments of Petitioner & 6.20 & 7.80 & 7.00 \\
Arguments of Respondent & 6.56 & 7.10 & 6.83 \\
Prediction & 6.88 & 7.60 & 7.24 \\
Explanation & 6.24 & 7.20 & 6.72 \\

\bottomrule
\end{tabular}
}

\caption{Expert evaluation of generated legal components across 25 sampled cases. Scores are reported on a 1--10 Likert scale.}

\label{tab:expert-scores}
\end{table}
%%%%%%%%%%%%%%%%%%%%%%%%%%%%%%%%%%
We evaluate the performance of \nm across multiple components of the CJPE pipeline, including legal issue identification, argument generation, prediction, and explanation. The results are reported using both automatic metrics and expert evaluation, as summarized in Table~\ref{tab:explanation-table}.

\noindent \textbf{Overall Performance:}  
Across all components, \texttt{Qwen3.5-27B} consistently outperforms \texttt{DeepSeek-R1-Distill-Qwen-14B (32-bit)} on most lexical and semantic metrics. In particular, improvements are observed in ROUGE, BLEU, and BERTScore across all reasoning components, indicating better alignment with ground truth and improved semantic consistency. Moreover, expert ratings also favor \texttt{Qwen3.5-27B}, demonstrating that improvements in automatic metrics translate into more coherent and legally meaningful outputs.

\noindent \textbf{Legal Issue and Argument Generation:}  
For intermediate reasoning components such as \textit{legal issue}, \textit{petitioner arguments}, and \textit{respondent arguments}, \texttt{Qwen3.5-27B} achieves higher scores across most evaluation metrics (Table~\ref{tab:explanation-table}). Notably, gains in ROUGE-2 and BERTScore indicate better capture of key legal phrases and improved contextual understanding. However, \texttt{DeepSeek-R1-Distill-Qwen-14B} achieves slightly higher METEOR scores in some cases, suggesting that it may generate more lexically varied outputs, albeit with slightly lower semantic alignment.

\noindent \textbf{Prediction Performance:}  
For the prediction task, \texttt{Qwen3.5-27B} achieves superior performance across all lexical and semantic metrics, with notable improvements in BLEU and ROUGE scores. This indicates that larger reasoning-oriented models are better at aligning their predictions with ground truth decisions when supported by structured reasoning and retrieved legal context. Expert evaluation further supports this observation, with \texttt{Qwen3.5-27B} achieving higher ratings for prediction quality.

\noindent \textbf{Explanation Quality:}  
In the explanation generation task, we observe a trade-off between lexical overlap and semantic richness. While \texttt{DeepSeek-R1-Distill-Qwen-14B} achieves slightly higher ROUGE scores, \texttt{Qwen3.5-27B} outperforms in METEOR, BERTScore, BLANC, and expert ratings. This suggests that \texttt{Qwen3.5-27B} generates explanations that are more semantically meaningful, coherent, and aligned with legal reasoning, even if they do not strictly match the reference text at the surface level.

\noindent \textbf{Expert Evaluation Analysis:}  
Table~\ref{tab:expert-scores} presents detailed expert evaluation across different components. \texttt{Qwen3.5-27B} consistently achieves higher average scores across all categories, particularly in \textit{prediction} (7.24) and \textit{legal issue} (7.10), indicating stronger reasoning capabilities and better alignment with judicial interpretation. The relatively lower scores for explanation (6.72) suggest that generating high-quality legal explanations remains a challenging task even for large models. In comparison, \texttt{DeepSeek-R1-Distill-Qwen-14B} produces moderately coherent outputs but lacks consistency in explanation quality.

\noindent \textbf{Impact of Quantization:}  
We observe a clear degradation in reasoning quality with aggressive quantization. The 4-bit variant of \texttt{DeepSeek-R1-Distill-Qwen-14B} frequently produces repetitive and semantically redundant outputs, indicating instability in generation. The 8-bit variant shows partial improvement but still struggles to adhere to structured reasoning templates. In contrast, higher precision variants (16-bit and 32-bit) generate more coherent, logically structured, and template-consistent outputs. This highlights the importance of maintaining sufficient numerical precision for tasks requiring multi-step reasoning and structured generation. Additional qualitative examples illustrating these effects are provided in Table~\ref{tab:quantization-effects} in the Appendix.

\noindent \textbf{Key Observations:}  
(i) Larger models demonstrate stronger capability in structured legal reasoning and explanation generation.  
(ii) Semantic metrics and expert evaluations provide more reliable insights than purely lexical metrics for legal tasks.  
(iii) Retrieval-grounded reasoning significantly improves both prediction accuracy and explanation quality.  
(iv) Quantization introduces a trade-off between efficiency and reasoning fidelity, with lower-bit models exhibiting reduced stability and coherence.

\vspace{0.5em}
\noindent Overall, the results demonstrate that \nm effectively enables structured, transparent, and evidence-grounded legal reasoning, with significant improvements observed when leveraging larger reasoning-oriented LLMs and higher-precision configurations.

\section{Conclusion and Future Work}

In this work, we presented \nmbf, a framework for Court Judgment Prediction and Explanation (CJPE) that integrates retrieval, structured reasoning, and prediction for the Indian legal domain. Unlike prior approaches that focus primarily on prediction or surface-level explanations, \nm explicitly models judicial reasoning by incorporating legal issues, arguments, and grounding in statutes and precedents. Our results demonstrate that combining RAG with reasoning-oriented LLMs improves semantic alignment, explanation quality, and expert-rated performance, while highlighting the importance of model scale and precision for stable reasoning.

Despite these contributions, the current system is limited to English-language data and does not fully enforce verification of reasoning against legal sources. Future work will focus on extending the framework to multilingual Indian legal settings, incorporating reinforcement learning and verification mechanisms to improve factual consistency, and exploring agentic or iterative reasoning for deeper legal analysis. These directions aim to further enhance the transparency, reliability, and practical applicability of Legal AI systems.

\section*{Limitations}

This study has several limitations that should be considered when interpreting the results. First, \nm is constrained by the context length of the underlying language models. In our current setup, the maximum input length is limited to 16,384 tokens and the output to 4,096 tokens. However, real-world legal documents often exceed these limits, typically ranging from 32K to 64K tokens. As a result, documents must be truncated or selectively processed, which may omit relevant information and affect the completeness of the generated reasoning and explanations.

Second, the framework is limited to English-language legal data. This is primarily due to the scarcity of large-scale, high-quality annotated datasets in Indian regional languages. Consequently, the applicability of the system remains restricted in multilingual judicial settings, particularly in district and subordinate courts where regional languages are widely used.

Finally, while \nm improves transparency through structured reasoning and retrieval grounding, it does not enforce strict verification of generated reasoning against legal sources. This may still lead to partially grounded or inconsistent explanations in complex cases.

Addressing these limitations, particularly improving long-context handling, enabling multilingual support, and incorporating stronger verification mechanisms, remains important for enhancing the robustness and real-world applicability of the proposed system.

% \section*{Limitation}
% This study has several limitations that should be acknowledged when interpreting the results. Nyayamind is constrained by the context length available to the language model. In the current implementation, the maximum input context is limited to 16,384 tokens, and the generated output is restricted to 4,096 tokens. However, real-world legal documents exceed these limits, frequently ranging between 32,000 and 64,000 tokens. Long documents must be truncated or selectively processed before being provided to the model. This may lead to the exclusion of potentially relevant information and could affect the completeness of the generated legal reasoning. Furthermore, complex legal cases may require extended chains of reasoning, which may exceed the current output token limit, thereby restricting the model’s ability to fully articulate detailed legal explanations.

% Nyayamind is limited to the English language. This limitation primarily arises from the lack of large scale, high quality legal datasets in Indian regional languages that are suitable for training and evaluation. This language constraint restricts the broader applicability of the system to district and lower courts, where legal proceedings and documentation often occur in regional languages.

% Addressing these limitations—particularly expanding context capacity and incorporating multilingual legal datasets—will be essential steps toward improving the scalability and real-world applicability of the proposed system.

\section*{Ethical Considerations}

The development and deployment of \nm in the legal domain raises several important ethical considerations. First, legal decision-making is a high-stakes process, and any AI-assisted system must be used with caution. \nm is designed as a decision-support tool and not as a replacement for human judges or legal professionals. The outputs generated by the system should be interpreted as advisory and must always be reviewed by qualified legal experts before use in real-world settings.

Second, the quality and bias of training data can influence model behavior. Although our datasets are sourced from publicly available legal documents, they may reflect historical biases present in judicial decisions. As a result, the model may inadvertently learn and propagate such biases. Addressing fairness and bias mitigation in legal AI systems remains an important direction for future work.

Third, while \nm improves transparency through structured reasoning and retrieval grounding, large language models may still generate hallucinated or partially incorrect information. This could lead to misleading interpretations if not carefully validated. To mitigate this, we emphasize the importance of grounding outputs in retrieved legal sources and recommend incorporating additional verification mechanisms in future iterations.

Fourth, privacy and data sensitivity are important considerations. Although the datasets used in this work consist of publicly available judgments and statutes, care must be taken when extending the system to other domains that may involve sensitive or personal information.

Finally, the deployment of Legal AI systems should align with regulatory and ethical frameworks governing AI usage. Ensuring accountability, transparency, and human oversight is critical for building trustworthy AI systems in the legal domain. We advocate for responsible use of such technologies, with clear boundaries on their role in supporting, not replacing, judicial processes.
\bibliography{custom,legal_bib}

@article{papadouli2022transparency,
  title={Transparency in artificial intelligence: A legal perspective},
  author={Papadouli, Vasiliki},
  journal={Journal of Ethics and Legal Technologies},
  volume={4},
  pages={25--40},
  year={2022}
}

@article{benedetto2025boosting,
  title={Boosting court judgment prediction and explanation using legal entities: I. Benedetto et al.},
  author={Benedetto, Irene and Koudounas, Alkis and Vaiani, Lorenzo and Pastor, Eliana and Cagliero, Luca and Tarasconi, Francesco and Baralis, Elena},
  journal={Artificial Intelligence and Law},
  volume={33},
  number={3},
  pages={605--640},
  year={2025},
  publisher={Springer}
}

@inproceedings{nigam-etal-2024-rethinking,
    title = "Rethinking Legal Judgement Prediction in a Realistic Scenario in the Era of Large Language Models",
    author = "Nigam, Shubham Kumar  and
      Deroy, Aniket  and
      Maity, Subhankar  and
      Bhattacharya, Arnab",
    editor = "Aletras, Nikolaos  and
      Chalkidis, Ilias  and
      Barrett, Leslie  and
      Goanț{\u{a}}, C{\u{a}}t{\u{a}}lina  and
      Preoțiuc-Pietro, Daniel  and
      Spanakis, Gerasimos",
    booktitle = "Proceedings of the Natural Legal Language Processing Workshop 2024",
    month = nov,
    year = "2024",
    address = "Miami, FL, USA",
    publisher = "Association for Computational Linguistics",
    url = "https://aclanthology.org/2024.nllp-1.6/",
    doi = "10.18653/v1/2024.nllp-1.6",
    pages = "61--80"
}

@article{huang2021dependency,
  title={Dependency learning for legal judgment prediction with a unified text-to-text transformer},
  author={Huang, Yunyun and Shen, Xiaoyu and Li, Chuanyi and Ge, Jidong and Luo, Bin},
  journal={arXiv preprint arXiv:2112.06370},
  year={2021}
}

@ARTICLE{10750819,
  author={Wang, Xuran and Zhang, Xinguang and Hoo, Vanessa and Shao, Zhouhang and Zhang, Xuguang},
  journal={IEEE Access}, 
  title={LegalReasoner: A Multi-Stage Framework for Legal Judgment Prediction via Large Language Models and Knowledge Integration}, 
  year={2024},
  volume={12},
  number={},
  pages={166843-166854},
  doi={10.1109/ACCESS.2024.3496666}}

@inproceedings{nigam-etal-2024-legal,
    title = "Legal Judgment Reimagined: {P}red{E}x and the Rise of Intelligent {AI} Interpretation in {I}ndian Courts",
    author = "Nigam, Shubham Kumar  and
      Sharma, Anurag  and
      Khanna, Danush  and
      Shallum, Noel  and
      Ghosh, Kripabandhu  and
      Bhattacharya, Arnab",
    editor = "Ku, Lun-Wei  and
      Martins, Andre  and
      Srikumar, Vivek",
    booktitle = "Findings of the Association for Computational Linguistics: ACL 2024",
    month = aug,
    year = "2024",
    address = "Bangkok, Thailand",
    publisher = "Association for Computational Linguistics",
    url = "https://aclanthology.org/2024.findings-acl.255/",
    doi = "10.18653/v1/2024.findings-acl.255",
    pages = "4296--4315"
}

@inproceedings{luo-etal-2023-prototype,
    title = "Prototype-Based Interpretability for Legal Citation Prediction",
    author = "Luo, Chu Fei  and
      Bhambhoria, Rohan  and
      Dahan, Samuel  and
      Zhu, Xiaodan",
    editor = "Rogers, Anna  and
      Boyd-Graber, Jordan  and
      Okazaki, Naoaki",
    booktitle = "Findings of the Association for Computational Linguistics: ACL 2023",
    month = jul,
    year = "2023",
    address = "Toronto, Canada",
    publisher = "Association for Computational Linguistics",
    url = "https://aclanthology.org/2023.findings-acl.301/",
    doi = "10.18653/v1/2023.findings-acl.301",
    pages = "4883--4898"
}

@inproceedings{wu-etal-2023-precedent,
    title = "Precedent-Enhanced Legal Judgment Prediction with {LLM} and Domain-Model Collaboration",
    author = "Wu, Yiquan  and
      Zhou, Siying  and
      Liu, Yifei  and
      Lu, Weiming  and
      Liu, Xiaozhong  and
      Zhang, Yating  and
      Sun, Changlong  and
      Wu, Fei  and
      Kuang, Kun",
    editor = "Bouamor, Houda  and
      Pino, Juan  and
      Bali, Kalika",
    booktitle = "Proceedings of the 2023 Conference on Empirical Methods in Natural Language Processing",
    month = dec,
    year = "2023",
    address = "Singapore",
    publisher = "Association for Computational Linguistics",
    url = "https://aclanthology.org/2023.emnlp-main.740/",
    doi = "10.18653/v1/2023.emnlp-main.740",
    pages = "12060--12075"
}

@article{makauskaite2025transparency,
  title={Transparency in the Labyrinths of the EU AI Act: Smart or Disbalanced?},
  author={Makauskaite-Samuole, Gintare},
  journal={Access to Just. E. Eur.},
  pages={38},
  year={2025},
  publisher={HeinOnline}
}

@inproceedings{deng-etal-2024-enabling,
    title = "Enabling Discriminative Reasoning in {LLM}s for Legal Judgment Prediction",
    author = "Deng, Chenlong  and
      Mao, Kelong  and
      Zhang, Yuyao  and
      Dou, Zhicheng",
    editor = "Al-Onaizan, Yaser  and
      Bansal, Mohit  and
      Chen, Yun-Nung",
    booktitle = "Findings of the Association for Computational Linguistics: EMNLP 2024",
    month = nov,
    year = "2024",
    address = "Miami, Florida, USA",
    publisher = "Association for Computational Linguistics",
    url = "https://aclanthology.org/2024.findings-emnlp.43/",
    doi = "10.18653/v1/2024.findings-emnlp.43",
    pages = "784--796"
}

@inproceedings{gan2021judgment,
  title={Judgment prediction via injecting legal knowledge into neural networks},
  author={Gan, Leilei and Kuang, Kun and Yang, Yi and Wu, Fei},
  booktitle={Proceedings of the AAAI conference on artificial intelligence},
  volume={35},
  pages={12866--12874},
  year={2021}
}

@article{esposito2022transparency,
  title={Transparency versus explanation: The role of ambiguity in legal AI},
  author={Esposito, Elena},
  journal={Journal of Cross-disciplinary Research in Computational Law},
  volume={1},
  number={2},
  year={2022}
}

@inproceedings{10.1145/1047788.1047838,
author = {Bruninghaus, Stefanie and Ashley, Kevin D.},
title = {Predicting outcomes of case based legal arguments},
year = {2003},
isbn = {1581137478},
publisher = {Association for Computing Machinery},
address = {New York, NY, USA},
url = {https://doi.org/10.1145/1047788.1047838},
doi = {10.1145/1047788.1047838},
booktitle = {Proceedings of the 9th International Conference on Artificial Intelligence and Law},
pages = {233–242},
numpages = {10},
location = {Scotland, United Kingdom},
series = {ICAIL '03}
}

@inproceedings{nigam-etal-2025-legalseg,
    title = "{L}egal{S}eg: Unlocking the Structure of {I}ndian Legal Judgments Through Rhetorical Role Classification",
    author = "Nigam, Shubham Kumar  and
      Dubey, Tanmay  and
      Sharma, Govind  and
      Shallum, Noel  and
      Ghosh, Kripabandhu  and
      Bhattacharya, Arnab",
    editor = "Chiruzzo, Luis  and
      Ritter, Alan  and
      Wang, Lu",
    booktitle = "Findings of the Association for Computational Linguistics: NAACL 2025",
    month = apr,
    year = "2025",
    address = "Albuquerque, New Mexico",
    publisher = "Association for Computational Linguistics",
    url = "https://aclanthology.org/2025.findings-naacl.63/",
    doi = "10.18653/v1/2025.findings-naacl.63",
    pages = "1129--1144",
    ISBN = "979-8-89176-195-7"
}

@inproceedings{nigam-etal-2025-nyayaanumana,
    title = "{NyayaAnumana} and {INLegalLlama}: The Largest {I}ndian Legal Judgment Prediction Dataset and Specialized Language Model for Enhanced Decision Analysis",
    author = "Nigam, Shubham Kumar  and
      Balaramamahanthi, Deepak Patnaik  and
      Mishra, Shivam  and
      Shallum, Noel  and
      Ghosh, Kripabandhu  and
      Bhattacharya, Arnab",
    editor = "Rambow, Owen  and
      Wanner, Leo  and
      Apidianaki, Marianna  and
      Al-Khalifa, Hend  and
      Eugenio, Barbara Di  and
      Schockaert, Steven",
    booktitle = "Proceedings of the 31st International Conference on Computational Linguistics",
    month = jan,
    year = "2025",
    address = "Abu Dhabi, UAE",
    publisher = "Association for Computational Linguistics",
    url = "https://aclanthology.org/2025.coling-main.738/",
    pages = "11135--11160"
}

@article{fan2025lexam,
  title={Lexam: Benchmarking legal reasoning on 340 law exams},
  author={Fan, Yu and Ni, Jingwei and Merane, Jakob and Tian, Yang and Hermstr{\"u}wer, Yoan and Huang, Yinya and Akhtar, Mubashara and Salimbeni, Etienne and Geering, Florian and Dreyer, Oliver and others},
  journal={arXiv preprint arXiv:2505.12864},
  year={2025}
}

@inproceedings{yao-etal-2025-elevating,
    title = "Elevating Legal {LLM} Responses: Harnessing Trainable Logical Structures and Semantic Knowledge with Legal Reasoning",
    author = "Yao, Rujing  and
      Wu, Yang  and
      Wang, Chenghao  and
      Xiong, Jingwei  and
      Wang, Fang  and
      Liu, Xiaozhong",
    editor = "Chiruzzo, Luis  and
      Ritter, Alan  and
      Wang, Lu",
    booktitle = "Proceedings of the 2025 Conference of the Nations of the Americas Chapter of the Association for Computational Linguistics: Human Language Technologies (Volume 1: Long Papers)",
    month = apr,
    year = "2025",
    address = "Albuquerque, New Mexico",
    publisher = "Association for Computational Linguistics",
    url = "https://aclanthology.org/2025.naacl-long.290/",
    doi = "10.18653/v1/2025.naacl-long.290",
    pages = "5630--5642",
    ISBN = "979-8-89176-189-6"
}

@inproceedings{cai2025unilaw,
  title={Unilaw-r1: A large language model for legal reasoning with reinforcement learning and iterative inference},
  author={Cai, Hua and Zhao, Shuang and Zhang, Liang and Shen, Xuli and Xu, Qing and Shen, Weilin and Wen, Zihao and Ban, Tianke},
  booktitle={Proceedings of the 2025 Conference on Empirical Methods in Natural Language Processing},
  pages={18128--18142},
  year={2025}
}

@article{yang2026lawthinker,
  title={LawThinker: A Deep Research Legal Agent in Dynamic Environments},
  author={Yang, Xinyu and Deng, Chenlong and Wen, Tongyu and Xie, Binyu and Dou, Zhicheng},
  journal={arXiv preprint arXiv:2602.12056},
  year={2026}
}

@article{zhou2026lras,
  title={LRAS: Advanced Legal Reasoning with Agentic Search},
  author={Zhou, Yujin and Cao, Chuxue and Yang, Jinluan and Wu, Lijun and He, Conghui and Han, Sirui and Guo, Yike},
  journal={arXiv preprint arXiv:2601.07296},
  year={2026}
}

@inproceedings{lin-2004-rouge,
    title = "{ROUGE}: A Package for Automatic Evaluation of Summaries",
    author = "Lin, Chin-Yew",
    booktitle = "Text Summarization Branches Out",
    month = jul,
    year = "2004",
    address = "Barcelona, Spain",
    publisher = "Association for Computational Linguistics",
    url = "https://aclanthology.org/W04-1013/",
    pages = "74--81"
}

@inproceedings{papineni2002bleu,
  title={Bleu: a method for automatic evaluation of machine translation},
  author={Papineni, Kishore and Roukos, Salim and Ward, Todd and Zhu, Wei-Jing},
  booktitle={Proceedings of the 40th annual meeting of the Association for Computational Linguistics},
  pages={311--318},
  year={2002}
}

@inproceedings{banerjee-lavie-2005-meteor,
    title = "{METEOR}: An Automatic Metric for {MT} Evaluation with Improved Correlation with Human Judgments",
    author = "Banerjee, Satanjeev  and
      Lavie, Alon",
    editor = "Goldstein, Jade  and
      Lavie, Alon  and
      Lin, Chin-Yew  and
      Voss, Clare",
    booktitle = "Proceedings of the {ACL} Workshop on Intrinsic and Extrinsic Evaluation Measures for Machine Translation and/or Summarization",
    month = jun,
    year = "2005",
    address = "Ann Arbor, Michigan",
    publisher = "Association for Computational Linguistics",
    url = "https://aclanthology.org/W05-0909/",
    pages = "65--72"
}

@misc{dettmers2023qloraefficientfinetuningquantized,
      title={QLoRA: Efficient Finetuning of Quantized LLMs}, 
      author={Tim Dettmers and Artidoro Pagnoni and Ari Holtzman and Luke Zettlemoyer},
      year={2023},
      eprint={2305.14314},
      archivePrefix={arXiv},
      primaryClass={cs.LG},
      url={https://arxiv.org/abs/2305.14314}, 
}

@misc{hu2021loralowrankadaptationlarge,
      title={LoRA: Low-Rank Adaptation of Large Language Models}, 
      author={Edward J. Hu and Yelong Shen and Phillip Wallis and Zeyuan Allen-Zhu and Yuanzhi Li and Shean Wang and Lu Wang and Weizhu Chen},
      year={2021},
      eprint={2106.09685},
      archivePrefix={arXiv},
      primaryClass={cs.CL},
      url={https://arxiv.org/abs/2106.09685}, 
}

@inproceedings{BERTScore,
  author       = {Tianyi Zhang and
                  Varsha Kishore and
                  Felix Wu and
                  Kilian Q. Weinberger and
                  Yoav Artzi},
  title        = {BERTScore: Evaluating Text Generation with {BERT}},
  booktitle    = {8th International Conference on Learning Representations, {ICLR} 2020,
                  Addis Ababa, Ethiopia, April 26-30, 2020},
  publisher    = {OpenReview.net},
  year         = {2020},
  url          = {https://openreview.net/forum?id=SkeHuCVFDr},
  bibsource    = {dblp computer science bibliography, https://dblp.org}
}

@article{blanc,
  author       = {Oleg V. Vasilyev and
                  Vedant Dharnidharka and
                  John Bohannon},
  title        = {Fill in the {BLANC:} Human-free quality estimation of document summaries},
  journal      = {CoRR},
  volume       = {abs/2002.09836},
  year         = {2020},
  url          = {https://arxiv.org/abs/2002.09836},
  eprinttype    = {arXiv},
  eprint       = {2002.09836},
  bibsource    = {dblp computer science bibliography, https://dblp.org}
}
\newpage
\appendix
\section{Implementation Details}
\label{sec:implementation}
This section provides implementation details of the retrieval and training components used in \nm. We describe three retrieval architectures, Endee, Milvus, and Vespa, along with their indexing strategies, hybrid retrieval mechanisms, and system configurations. Additionally, we present training and inference settings for the reasoning-oriented LLMs, as well as dataset statistics to support reproducibility.

\subsection{Dense Retrieval with Endee}

To facilitate effective semantic retrieval of legal documents, we utilized Endee for storing and querying dense vector representations. Each record included a unique string identifier, a 768-dimensional embedding created with the snowflake/snowflake-arctic-embed-m-v2.0, and the original document text (capped at 60,000 characters) along with structured metadata fields. An HNSW (Hierarchical Navigable Small World) index was set up with M = 16, ef\_construction = 128, and cosine similarity as the distance metric, employing FP16 (medium) storage precision to optimize both accuracy and memory usage. Documents underwent preprocessing through Unicode normalization, conversion to lowercase, and whitespace consolidation, followed by chunking with a RecursiveCharacterTextSplitter using a chunk size of 4,096 tokens and an overlap of 100 tokens, with the model's tokenizer serving as the length function. During query execution, the input text was encoded using the same snowflake/snowflake-arctic-embed-m-v2.0, and the resulting vector was compared against the HNSW index with ef = 128 and top\_k = 10, retrieving the top\_k document chunks that were most semantically similar based on cosine similarity.

This RAG setup is composed of four distinct collections, which are as follows: SC\_Judgments\_DB, Central\_Acts, State\_Acts, HC\_Judgments\_DB. The Central\_Acts collection comprises 36,991 segments. The HC\_Judgments\_DB collection holds 69,017 segments, while the SC\_Judgments\_DB collection contains 81,331 segments. The State\_Acts collection includes 1,50,748 segments. In addition to the embedding vector, these collections also retain metadata information. The schema of each collection is detailed in table \ref{tab:endee-indices}.

%%%%%%%%%%%%%%%%%%%%%%%%%%%%%%%%%%%%%%%%%%%%%%%
\begin{table}[t]
\centering
\small
\setlength{\tabcolsep}{4pt}
\renewcommand{\arraystretch}{1.12}

\begin{tabular}{l p{0.55\columnwidth}}
\toprule
\textbf{Parameter} & \textbf{Value} \\
\midrule

\multicolumn{2}{c}{\textbf{System}} \\
\addlinespace
GPUs & NVIDIA H200 \\

\midrule

\multicolumn{2}{c}{\textbf{LoRA Parameters}} \\
\addlinespace
Rank & 16 \\
Alpha & 16 \\
Dropout & 0 \\
Target Modules & q, k, v, o, gate, up, down projections \\

\midrule

\multicolumn{2}{c}{\textbf{Training Parameters}} \\
\addlinespace
Epochs & 8 \\
Batch Size & 1 \\
Max Sequence Length & 16,384 \\
Learning Rate & $1\times10^{-4}$ \\
Scheduler & Cosine annealing \\
Warmup Ratio & 0.1 \\
Weight Decay & 0.01 \\
Early Stopping Patience & 3 \\
Optimizer & AdamW (8-bit) \\

\midrule

\multicolumn{2}{c}{\textbf{Inference Parameters}} \\
\addlinespace
Max New Tokens & 4,096 \\
No Repeat N-gram Size & 6 \\
Repetition Penalty & 1.1 \\

\bottomrule
\end{tabular}

\caption{Training and inference hyperparameters used for the \textsc{Nyayamind} system.}

\label{tab:training-parameters}
\end{table}

\subsection{Hybrid Retrieval with Milvus}

We implemented Milvus as a self-hosted vector database to store and query dense vector representations. Each collection included a unique integer identifier, a 768-dimensional embedding created using the Snowflake/snowflake-arctic-embed-m-v2.0 model, and the original document text along with structured metadata fields. For collections such as Central Acts and Supreme Court, an IVF\_FLAT index was set up with nlist = 2048 and L2 (Euclidean) distance, enabling fast approximate nearest neighbor searches by dividing vectors into 2048 Voronoi cells. For collections like High Courts and State Acts, a FLAT (brute-force) index was employed for precise searches. Documents were divided into chunks using a RecursiveCharacterTextSplitter with a chunk size of 4,096 tokens and an overlap of 100 tokens, utilizing the model's tokenizer as the length function. During query time, the input was initially encoded and matched against the Milvus index with nprobe = 10, retrieving the top\_k = 500 semantically relevant candidates. BM25 was then selectively applied only to this vector pre-filtered subset, not the entire corpus, to capture lexical overlap while managing computational costs. The BM25 implementation used the Okapi BM25 formula with parameters b = 0.7 and k1 = 1.6, built on sklearn's TfidfVectorizer, and applied Porter stemming during tokenization. Documents were re-chunked into 512-character segments with a 100-character overlap for BM25 scoring. The ranked outputs from vector search and BM25 were combined using Reciprocal Rank Fusion (RRF) with a smoothing constant k = 60, resulting in a unified ranking that integrates both semantic similarity and lexical relevance signals. This three-stage approach — vector retrieval, subset BM25, and RRF fusion, balances retrieval accuracy with computational efficiency.

The RAG configuration comprises four main legal categories indexed in the Milvus vector database: Central Acts, Supreme Court Judgments, High Court Judgments, and State Acts, which together encompass 60 collections. The Central Acts category includes a single collection with 37,202 indexed segments. The Supreme Court category also has one collection, containing 81,625 segments. The High Courts category is divided state-wise into 25 distinct collections, totaling 1,577,859 indexed segments. The State Acts category consists of 33 state-wise collections with 150,878 indexed segments. Overall, the Milvus vector database holds 1,847,564 legal text segments across all collections. Each segment is stored with its corresponding embedding vector and related metadata, such as court name, act title, citation details, jurisdiction, and section identifiers where applicable. The schema design for each collection is outlined in Table \ref{tab:milvus-indices}.

\subsection{Parallel Hybrid Retrieval with Vespa}
The Vespa architecture implements a native hybrid approach where semantic and lexical searches operate in parallel across the entire corpus. We implemented Vespa as a self-hosted search engine within a Docker container. Each document schema included an HNSW-indexed 768-dimensional embedding, created using the snowflake/snowflake-arctic-embed-m-v2.0 model, alongside BM25-indexed text fields and structured metadata. The HNSW index was set up with a Euclidean distance metric, max\_links\_per\_node = 16, and neighbors\_to\_explore\_at\_insert = 200, utilizing paged attribute storage for efficient memory use. Documents were divided into chunks using a RecursiveCharacterTextSplitter with a chunk size of 8,192 tokens and an overlap of 100–200 tokens, using the model's tokenizer as the length function. Each schema specified two declarative rank profiles: a semantic profile employing `closeness(field, embedding)` for ANN-based nearest neighbor retrieval, and a BM25 profile for lexical scoring on text fields. During query time, the input was sent as two separate Vespa queries — a vector-based nearest neighbor search and a BM25 keyword search, each providing the top-10 results. The two ranked result lists were then combined client-side using Reciprocal Rank Fusion (RRF) with a smoothing constant of k = 60, resulting in a unified ranking that integrates both semantic similarity and lexical relevance signals. Unlike the Milvus-based architecture, BM25 was applied natively across the entire Vespa corpus in parallel with vector search rather than on a pre-filtered subset, and no cross-encoder re-ranking was used, leading to a two-stage retrieval pipeline.

The RAG setup is structured into four main legal categories: Central Acts, Supreme Court Judgments, High Court Judgments, and State Acts, all integrated under the legalknowledgebase application. The Central Acts category is represented by a single schema (centralacts) that includes 37,202 indexed segments. The Supreme Court category is managed through the scjudgments schema, which contains 81,625 indexed segments. The High Court collection is housed within a specific judgments schema, comprising 1,577,859 indexed segments sourced from 25 High Courts. The State Acts category is organized by state into 33 distinct schemas (sa\_{state}), together holding 150,878 indexed segments. The schema of each collection is detailed in table \ref{tab:vespa-indices}

The similarities and differences between the 3 RAG approaches are outlined in Table \ref{tab:rag-comparison-wide}.

%  Dataset Tokens
%%%%%%%%%%%%%%%%%%%%%%%%%%%%%%%%%%%%%%%%%%%%%%%
\begin{table*}[t]
\centering
\small
\setlength{\tabcolsep}{5pt}
\renewcommand{\arraystretch}{1.12}

\begin{tabular}{lrrrrrrrr}
\toprule

& \multicolumn{8}{c}{\textbf{Token Count Percentiles}} \\
\cmidrule(lr){2-9}

\textbf{Dataset Field} 
& \textbf{P50} 
& \textbf{P75} 
& \textbf{P80} 
& \textbf{P85} 
& \textbf{P90} 
& \textbf{P95} 
& \textbf{P99} 
& \textbf{Max} \\

\midrule
\multicolumn{9}{c}{\textbf{Train Dataset}} \\
\midrule

Input Text & 18,295 & 68,606 & 88,391 & 118,401 & 175,102 & 308,669 & 828,490 & 3,117,110 \\
Target Text & 2,096 & 3,274 & 3,747 & 4,407 & 5,485 & 8,289 & 15,243 & 33,839 \\
Case Text & 5,724 & 9,738 & 11,081 & 13,168 & 16,468 & 24,478 & 56,548 & 304,724 \\
Simplified Facts & 992 & 1,561 & 1,744 & 1,994 & 2,383 & 3,084 & 6,528 & 18,000 \\
Simplified Legal Issue & 61 & 116 & 143 & 173 & 218 & 295 & 563 & 8,724 \\
Petitioner Arguments & 360 & 678 & 780 & 946 & 1,235 & 1,679 & 3,993 & 13,970 \\
Respondent Arguments & 259 & 496 & 594 & 732 & 977 & 1,425 & 2,912 & 10,491 \\
Simplified Decision & 96 & 177 & 204 & 244 & 320 & 479 & 1,196 & 9,211 \\
Simplified Reasoning & 1,061 & 1,763 & 2,010 & 2,344 & 3,010 & 4,641 & 8,753 & 19,968 \\

\midrule
\multicolumn{9}{c}{\textbf{Validation Dataset}} \\
\midrule

Input Text & 17,927 & 61,311 & 78,422 & 109,823 & 177,678 & 282,347 & 1,279,851 & 2,700,242 \\
Target Text & 2,091 & 3,318 & 3,878 & 4,530 & 5,893 & 9,467 & 19,056 & 27,366 \\
Case Text & 5,888 & 9,604 & 11,043 & 12,814 & 16,328 & 25,305 & 68,429 & 413,279 \\
Simplified Facts & 976 & 1,535 & 1,679 & 1,967 & 2,337 & 2,839 & 5,020 & 11,689 \\
Simplified Legal Issue & 64 & 118 & 144 & 187 & 234 & 313 & 782 & 8,678 \\
Petitioner Arguments & 358 & 656 & 792 & 953 & 1,269 & 1,909 & 4,881 & 8,433 \\
Respondent Arguments & 260 & 534 & 634 & 822 & 1,032 & 1,628 & 2,696 & 9,799 \\
Simplified Decision & 100 & 183 & 227 & 279 & 363 & 547 & 828 & 2,877 \\
Simplified Reasoning & 1,032 & 1,774 & 2,043 & 2,345 & 3,027 & 5,010 & 12,303 & 17,994 \\

\midrule
\multicolumn{9}{c}{\textbf{Test Dataset}} \\
\midrule

Input Text & 18,992 & 73,468 & 95,149 & 125,702 & 186,462 & 328,597 & 865,328 & 2,133,870 \\
Target Text & 2,202 & 3,438 & 3,903 & 4,565 & 5,764 & 8,118 & 13,846 & 29,737 \\
Case Text & 5,982 & 9,933 & 11,247 & 13,010 & 15,893 & 21,284 & 46,887 & 153,313 \\
Simplified Facts & 996 & 1,531 & 1,754 & 2,024 & 2,386 & 3,222 & 8,549 & 11,837 \\
Simplified Legal Issue & 60 & 113 & 139 & 181 & 230 & 309 & 557 & 9,422 \\
Petitioner Arguments & 362 & 667 & 783 & 959 & 1,148 & 1,615 & 3,497 & 9,024 \\
Respondent Arguments & 264 & 504 & 613 & 781 & 1,035 & 1,500 & 3,234 & 10,122 \\
Simplified Decision & 95 & 173 & 202 & 240 & 298 & 443 & 824 & 2,035 \\
Simplified Reasoning & 1,158 & 1,904 & 2,135 & 2,515 & 3,182 & 4,608 & 8,514 & 19,802 \\

\bottomrule
\end{tabular}

\caption{Token distribution statistics across dataset fields. Values represent percentile statistics for token counts in the training, validation, and test splits.}

\label{tab:dataset-tokens}
\end{table*}

%  Endee Collections
%%%%%%%%%%%%%%%%%%%%%%%%%%%%%%%%%%%%%%%%%%%%%%%

\begin{table*}[t]
\centering
\small
\setlength{\tabcolsep}{6pt}
\renewcommand{\arraystretch}{1.15}

\begin{tabular}{lccp{8.5cm}}
\toprule

\textbf{Collection} 
& \textbf{\#Chunks} 
& \textbf{\#Collections} 
& \textbf{Metadata \& Filterable Fields} \\

\midrule

\texttt{Central\_Acts}
& 36,991
& 1
& \textbf{Metadata:} id, text, case\_name, section\_no, section\_title \\
&&&
\textbf{Filters:} section\_no, title, doc\_id \\

\addlinespace

\texttt{HC\_Judgments}
& 1,069,017
& 1
& \textbf{Metadata:} id, text, court, judge, CNR, date, disposal, state, case\_no \\
&&&
\textbf{Filters:} court, case\_no, state, doc\_id \\

\addlinespace

\texttt{SC\_Judgments}
& 81,331
& 1
& \textbf{Metadata:} id, text, case\_name, diary\_no, type, petitioner, respondent, bench, citations \\
&&&
\textbf{Filters:} case\_no, date, bench, case\_name \\

\addlinespace

\texttt{State\_Acts}
& 150,748
& 1
& \textbf{Metadata:} id, text, state\_name, statute, section\_no, section\_title \\
&&&
\textbf{Filters:} state, statute, section\_no \\

\bottomrule
\end{tabular}

\caption{Collections stored in the Endee vector database used for legal retrieval. Each collection contains chunked legal documents with associated metadata fields and filterable attributes enabling structured retrieval.}

\label{tab:endee-indices}
\end{table*}

%  Vespa Collections
%%%%%%%%%%%%%%%%%%%%%%%%%%%%%%%%%%%%%%%%%%%%%%%
\begin{table*}[t]
\centering
\small
\setlength{\tabcolsep}{6pt}
\renewcommand{\arraystretch}{1.15}

\begin{tabular}{lccp{8.5cm}}
\toprule

\textbf{Collection} 
& \textbf{\#Chunks} 
& \textbf{\#Collections} 
& \textbf{BM25 Fields \& Metadata} \\

\midrule

\texttt{centralacts}
& 36,921
& 1
& \textbf{BM25 Fields:} section\_text, section\_title, statute \\
&&&
\textbf{Metadata:} id, section\_no \\

\addlinespace

\texttt{scjudgments}
& 52,796
& 1
& \textbf{BM25 Fields:} content \\
&&&
\textbf{Metadata:} case\_name, diary\_no, type, case\_no, petitioner, respondent, bench, date, citations \\

\addlinespace

\texttt{hcjudgments}
& 811,492
& 1
& \textbf{BM25 Fields:} text, case\_name \\
&&&
\textbf{Metadata:} id, judgment\_date, citations, bench \\

\addlinespace

\texttt{sa\_\{state\}}
& 50,156
& 33
& \textbf{BM25 Fields:} section\_text, section\_title \\
&&&
\textbf{Metadata:} id, state, statute, section\_no, doc\_id \\

\bottomrule
\end{tabular}

\caption{Collections stored in the Vespa vector database. Each collection defines searchable BM25 fields along with structured metadata attributes used for filtering and retrieval.}

\label{tab:vespa-indices}
\end{table*}

%  Milvus Collections
%%%%%%%%%%%%%%%%%%%%%%%%%%%%%%%%%%%%%%%%%%%%%%%
\begin{table*}[t]
\centering
\small
\setlength{\tabcolsep}{6pt}
\renewcommand{\arraystretch}{1.15}

\begin{tabular}{lccp{8.5cm}}
\toprule

\textbf{Collection} 
& \textbf{\#Chunks} 
& \textbf{\#Collections} 
& \textbf{Metadata Fields} \\

\midrule

\texttt{Central\_Acts}
& 37,202
& 1
& id, embedding, text, case\_name, section\_no, section\_title \\

\addlinespace

\texttt{SC\_Judgments}
& 81,625
& 1
& id, embedding, text, case\_name, diary\_no, type, petitioner, respondent, bench, citations \\

\addlinespace

\texttt{HC\_\{Court\}}
& 1,577,838
& 25
& id, embedding, case\_no, pdf\_path, state, disposal, decision\_date, text, title, judge, CNR \\

\addlinespace

\texttt{State\_Acts\_\{Name\}}
& 146,804
& 33
& id, embedding, doc\_id, state\_name, statute\_name, section\_no, section\_title, text \\

\bottomrule
\end{tabular}

\caption{Collections stored in the Milvus vector database. Each collection stores embedded legal text chunks along with structured metadata attributes used for filtering and retrieval.}

\label{tab:milvus-indices}
\end{table*}

%%%%%%%%%%%%%%%%%%%%%%%%%%%%%%%%%%%%%%%%%%%%%%%
% 3 RAG Architectures(pasted from methodology.tex)
%%%%%%%%%%%%%%%%%%%%%%%%%%%%%%%%%%%%%%%%%%%%%%%

\begin{table*}[t]
\centering
\small
\setlength{\tabcolsep}{6pt}
\renewcommand{\arraystretch}{1.2}

\resizebox{\textwidth}{!}{
\begin{tabular}{lccc}
\toprule
\textbf{Feature} & \textbf{Endee} & \textbf{Milvus} & \textbf{VespaDB} \\
\midrule

\multicolumn{4}{c}{\textbf{Retrieval Strategy}} \\
\midrule

Retrieval Philosophy
& \makecell{Single-stage \\ Dense Retrieval}
& \makecell{Three-stage Hybrid \\ Semantic $\rightarrow$ Lexical $\rightarrow$ Fusion}
& \makecell{Parallel Hybrid \\ (Semantic + Lexical) $\rightarrow$ Fusion} \\

Fusion Method
& --
& RRF ($k=60$)
& RRF ($k=60$) \\

BM25 Retrieval
& Not used
& \makecell{Applied on \\ top-500 vector results}
& \makecell{Native parallel \\ full-corpus BM25} \\

\midrule

\multicolumn{4}{c}{\textbf{Vector Configuration}} \\
\midrule

Embedding Model
& \multicolumn{3}{c}{\texttt{snowflake-arctic-embed-m-v2.0} (768-d)} \\

Distance Metric
& Cosine & L2 & L2 \\

Index Type
& \makecell{HNSW \\ (M=16, ef\_con=128)}
& \makecell{IVF\_FLAT \\ (nlist=2048) or FLAT}
& \makecell{HNSW \\ (M=16, ef\_ins=200)} \\

Precision
& FP16
& FP32
& \makecell{FP32 \\ (Paged attributes)} \\

Chunk Size
& 4096 & 4096 & 8192 \\

\bottomrule
\end{tabular}
}

\caption{Comparison of the three retrieval architectures used in \textsc{Nyayamind}. The systems differ in retrieval strategy, hybrid fusion mechanisms, and vector indexing configurations.}
\label{tab:rag-comparison-wide}
\end{table*}

%%%%%%%%%%%%%%%%%%%%%%%%%%%%%%%%%%%%%%%%%%%%%%%
\begin{table*}[ht]
%\small
    \centering
    \begin{tabular}{|p{0.8\textwidth}|}
    \hline
{\bf Input Text}\\
\hline
   
{\bf System Message} =``SYSTEM:\\ You are a smart and intelligent legal assistant for the Indian legal domain. Based on the user's instructions, you will have to perform or assist in some tasks related to the Indian legal system. Since these tasks have some legal application, only provide responses you are extremely certain about, and avoid being ambiguous or uncertain. Ensure that your outputs adhere to the user's instructions or requirements.''\\

{\bf User Content} = ``USER: \\
    You are a legal expert tasked with making a judgment about whether an appeal should be accepted or rejected based on the provided case proceeding, cited statutes and cited cases. Your task is to evaluate whether the appeal should be accepted (1) or rejected (0) based on the input.

    \#\#\# Now, evaluate the following case: \\
    Case Proceedings: \textbf{\{simplified\_facts\}}
    
    Relevant Statutes:
    \textbf{\{sections\_texts\}}
    
    Cited Cases Reference:
    \textbf{\{cited\_case\_judgments\}}
    
    Provide your judgment by strictly following this format: \\
    \#\#PREDICTION: [Insert your prediction here] \\
    \#\#EXPLANATION: [Insert your reasoning here that led you to your prediction.] \\
    Strictly do not include anything outside this format. Strictly follow the provided format. Do not generate placeholders. Just provide the final judgment and explanation.'' \\
\hline
    {\bf Target Text}\\
    \hline
   
``ASSISTANT: \\ \#\#\# Response: \\ 
<think>\\ 
        ** Legal Issue Analysis:**
        \textbf{\{simplified\_issue\}}

        **Arguments of Petitioner:**
        \textbf{\{simplified\_arguments\_of\_petitioner\}}

        **Arguments of Respondent:**
        \textbf{\{simplified\_arguments\_of\_respondent\}}

        **Deliberation:**
        Weighing the arguments against the relevant statutes and cited cases to form a decision. \\</think> \\ 
        \#\#PREDICTION: \textbf{\{simplified\_decision\}} \\ \#\#EXPLANATION: \textbf{\{simplified\_reasoning\}}
''\\

\hline

    \end{tabular}
    \caption{Prompt for DeepSeek-R1-Distill-Qwen-14B model.}
    \label{tab:prompt-deepseek-r1}
\end{table*}
%%%%%%%%%%%%%%%%%%%%%%%%%%%%%%%%%%%%%%%%%%%%%%%

%%%%%%%%%%%%%%%%%%%%%%%%%%%%%%%%%%%%%%%%%%%%%%%
\begin{table*}[ht]
%\small
    \centering
    \begin{tabular}{|p{0.8\textwidth}|}
    \hline
{\bf Input Text}\\
\hline
   
{\bf System Message} =``<|system|> You are a smart and intelligent legal assistant for the Indian legal domain. Based on the user's instructions, you will have to perform or assist in some tasks related to the Indian legal system. Since these tasks have some legal application, only provide responses you are extremely certain about, and avoid being ambiguous or uncertain. Ensure that your outputs adhere to the user's instructions or requirements. \\<|end|>''\\

{\bf User Content} = ``<|user|> You are a legal expert tasked with making a judgment about whether an appeal should be accepted or rejected based on the provided case proceeding, cited statutes and cited cases. Your task is to evaluate whether the appeal should be accepted (1) or rejected (0) based on the input.\\
        \#\#\# Now, evaluate the following case:\\
        Case Proceedings: \textbf{\{simplified\_facts\}}\\
        Relevant Statutes:\\ \textbf{\{sections\_texts\}}\\
        Cited Cases Reference:\\ \textbf{\{cited\_case\_judgments\}}\\
        Provide your judgment by strictly following this format:\\
        \#\#PREDICTION: [Insert your prediction here]\\
        \#\#EXPLANATION: [Insert your reasoning here that led you to your prediction.]\\
        Strictly do not include anything outside this format. Strictly follow the provided format. Do not generate placeholders. Just provide the final judgment and explanation.<|end|>'' \\

{\bf Assistant} = ``<|assistant|>''\\      

\hline
    {\bf Target Text}\\
    \hline
   
``<think>\\**Legal Issue Analysis:**\\
        \textbf{\{simplified\_issue\}}\\
        **Arguments of Petitioner:**\\
        \textbf{\{simplified\_arguments\_of\_petitioner\}}\\
        **Arguments of Respondent:**\\
        \textbf{\{simplified\_arguments\_of\_respondent\}}\\
        **Deliberation:**\\
        Weighing the arguments against the relevant statutes and cited cases to form a decision.\\</think>\\ \#\#PREDICTION: \textbf{\{simplified\_decision\}} \\ \#\#EXPLANATION: \textbf{\{simplified\_reasoning\}}''\\
\hline
    \end{tabular}
    \caption{Prompt for Phi-4-mini-reasoning model.}
    \label{tab:prompt-phi4-mini-reasoning}
\end{table*}
%%%%%%%%%%%%%%%%%%%%%%%%%%%%%%%%%%%%%%%%%%%%%%%

%%%%%%%%%%%%%%%%%%%%%%%%%%%%%%%%%%%%%%%%%%%%%%%
\begin{table*}[ht]
%\small
    \centering
    \begin{tabular}{|p{0.8\textwidth}|}
    \hline
{\bf Input Text}\\
\hline
   
{\bf System Message} =``<|im\_start|>system<|im\_sep|> \\ You are a smart and intelligent legal assistant for the Indian legal domain. Based on the user's instructions, you will have to perform or assist in some tasks related to the Indian legal system. Since these tasks have some legal application, only provide responses you are extremely certain about, and avoid being ambiguous or uncertain. Ensure that your outputs adhere to the user's instructions or requirements. \\<|im\_end|>''\\

{\bf User Content} = ``<|im\_start|>user<|im\_sep|> \\ You are a legal expert tasked with making a judgment about whether an appeal should be accepted or rejected based on the provided case proceeding, cited statutes and cited cases. Your task is to evaluate whether the appeal should be accepted (1) or rejected (0) based on the input.\\
        \#\#\# Now, evaluate the following case:\\
        Case Proceedings: \textbf{\{simplified\_facts\}}\\
        Relevant Statutes:\\ \textbf{\{sections\_texts\}}\\
        Cited Cases Reference:\\ \textbf{\{cited\_case\_judgments\}}\\
        Provide your judgment by strictly following this format:\\
        \#\#PREDICTION: [Insert your prediction here]\\
        \#\#EXPLANATION: [Insert your reasoning here that led you to your prediction.]\\
        Strictly do not include anything outside this format. Strictly follow the provided format. Do not generate placeholders. Just provide the final judgment and explanation.<|im\_end|>'' \\
        {\bf Assistant} = ``<|im\_start|>assistant<|im\_sep|>''\\    
\hline
    {\bf Target Text}\\
    \hline
   
``<think>\\**Legal Issue Analysis:**\\
        \textbf{\{simplified\_issue\}}\\
        **Arguments of Petitioner:**\\
        \textbf{\{simplified\_arguments\_of\_petitioner\}}\\
        **Arguments of Respondent:**\\
        \textbf{\{simplified\_arguments\_of\_respondent\}}\\
        **Deliberation:**\\
        Weighing the arguments against the relevant statutes and cited cases to form a decision.\\</think>\\ \#\#PREDICTION: \textbf{\{simplified\_decision\}} \\ \#\#EXPLANATION: \textbf{\{simplified\_reasoning\}}''\\
\hline
    \end{tabular}
    \caption{Prompt for Phi-4-reasoning model.}
    \label{tab:prompt-phi4-reasoning}
\end{table*}
%%%%%%%%%%%%%%%%%%%%%%%%%%%%%%%%%%%%%%%%%%%%%%%

%%%%%%%%%%%%%%%%%%%%%%%%%%%%%%%%%%%%%%%%%%%%%%%
\begin{table*}[ht]
%\small
    \centering
    \begin{tabular}{|p{0.8\textwidth}|}
    \hline
{\bf Input Text}\\
\hline
   
{\bf System Message} =``<|im\_start|><|system|> \\ You are a smart and intelligent legal assistant for the Indian legal domain. Based on the user's instructions, you will have to perform or assist in some tasks related to the Indian legal system. Since these tasks have some legal application, only provide responses you are extremely certain about, and avoid being ambiguous or uncertain. Ensure that your outputs adhere to the user's instructions or requirements. \\<|im\_end|>''\\

{\bf User Content} = ``<|im\_start|><|user|> You are a legal expert tasked with making a judgment about whether an appeal should be accepted or rejected based on the provided case proceeding, cited statutes and cited cases. Your task is to evaluate whether the appeal should be accepted (1) or rejected (0) based on the input.\\
        \#\#\# Now, evaluate the following case:\\
        Case Proceedings: \textbf{\{simplified\_facts\}}\\
        Relevant Statutes:\\ \textbf{\{sections\_texts\}}\\
        Cited Cases Reference:\\ \textbf{\{cited\_case\_judgments\}}\\
        Provide your judgment by strictly following this format:\\
        \#\#PREDICTION: [Insert your prediction here]\\
        \#\#EXPLANATION: [Insert your reasoning here that led you to your prediction.]\\
        Strictly do not include anything outside this format. Strictly follow the provided format. Do not generate placeholders. Just provide the final judgment and explanation.<|im\_end|>
        '' \\
        {\bf Assistant} = ``<|im\_start|><|assistant|>''\\    
\hline
    {\bf Target Text}\\
    \hline
   
``<think>\\**Legal Issue Analysis:**\\
        \textbf{\{simplified\_issue\}}\\
        **Arguments of Petitioner:**\\
        \textbf{\{simplified\_arguments\_of\_petitioner\}}\\
        **Arguments of Respondent:**\\
        \textbf{\{simplified\_arguments\_of\_respondent\}}\\
        **Deliberation:**\\
        Weighing the arguments against the relevant statutes and cited cases to form a decision.\\</think>\\ \#\#PREDICTION: \textbf{\{simplified\_decision\}} \\ \#\#EXPLANATION: \textbf{\{simplified\_reasoning\}}''\\
\hline
    \end{tabular}
    \caption{Prompt for Qwen3.5-27B model.}
    \label{tab:prompt-qwen3.5}
\end{table*}
%%%%%%%%%%%%%%%%%%%%%%%%%%%%%%%%%%%%%%%%%%%%%%%
\begin{table*}[ht]
\centering
\small
\renewcommand{\arraystretch}{1.05}
\begin{tabularx}{\textwidth}{|p{3cm}|X|}

\hline

\multicolumn{1}{|c|}{\textbf{Model}} & \centering\arraybackslash \textbf{Reasoning, Decision and Explanation} \\ \hline
 
Ground Truth &
  { 
  \textbf{Legal Issue} \par
  Is the appellant considered to be in "possession" of the lands as khudkasht\dots \par
  \textbf{Arguments of Petitioner} \par
  It can be contended by Mr. J. P. Goyal, learned counsel for the appellant\dots \par
  \textbf{Arguments of Respondent} \par
  The respondent\'s arguments could be that the appellant may not be able to challenge the final\dots
 \textbf{Decision}\par
 We may hold that the judgment\dots \textbf{appeal can be set aside}. \par
 \textbf{Explanation}
 It can be inferred that the normal principle of possession by one co-sharer\dots\par
 } \\ \hline
 DeepSeek-R1-Distill-Qwen-14B (4-bit) &
  { 
  \textbf{**Legal Issue Analysis:**} \par  Whether the plaintiff was in cultivatory possession of the plots in dispute on the date of vesting? \par      \textbf{**Arguments of Petitioner:**}\par      It can be inferred that the petitioner may argue that the plaintiff's counsel\dots \par
  \textbf{**Arguments of Respondent:**} \par 
  The respondent may contend that the plaintiff's possession of the plots in dispute may have been limited to a mere physical presence \dots \par \colorbox{blue!10}{%
    \parbox{\dimexpr\linewidth-2\fboxsep}{%
    The respondent may argue that the plaintiff's possession of the plots in dispute may have been subject to the rights of the defendants\dots \textbf{(Sementic\_Redundancy)}}} \par
  \textbf{\#\#PREDICTION:} 16. The appeal may be dismissed. \par
  \textbf{\#\# EXPLANATION:} 
  It can be inferred that the plaintiff's possession of the plots \d in dispute was not cultivatory possession, as per the findings of the High Court.  \colorbox{red!10}{%
    \parbox{\dimexpr\linewidth-2\fboxsep}{%
    It can be inferred that the plaintiff's possession of the plots \dots \textbf{(Repetitive Text)}}}
  } \\ \hline

DeepSeek-R1-Distill-Qwen-14B (8-bit) &
  { 
  \colorbox{yellow!20}{%
    \parbox{\dimexpr\linewidth-2\fboxsep}{%
    Alright, I\'m trying to figure out whether the appeal in this case should be accepted or rejected. }} \par \colorbox{yellow!10}{%
    \parbox{\dimexpr\linewidth-2\fboxsep}{%
    First, looking at the case proceedings, the plaintiff, Kailashrai\dots\textbf{(Wrong output format)}}} \par \textbf{\#\#PREDICTION:} 0 \par \textbf{\#\#EXPLANATION:} \par The High Court correctly determined that the plaintiff was not in cultivatory\dots
  } \\ \hline
DeepSeek-R1-Distill-Qwen-14B (16-bit) &
  {
  \textbf{**Legal Issue Analysis:**}\par Is the plaintiff entitled to a declaration that he is entitled to one-fourth share in the properties? \dots \par \textbf{**Arguments of Petitioner:**} \par It is contended that the plaintiff can establish his right to cultivate the plots in question \dots \par \textbf{**Arguments of Respondent:**} \par The respondent's arguments could be that the plaintiff may not be able to establish his right \dots \par 
  \textbf{\#\#PREDICTION:} \colorbox{green!20}{ The appeal can be dismissed with costs\dots\textbf{(Correct Prediction)}}\par \textbf{\#\#EXPLANATION:} It can be inferred that the plaintiff's claim is based on the  \dots
  } \\ \hline
  
DeepSeek-R1-Distill-Qwen-14B (32-bit) &
  {
  \textbf{**Legal Issue Analysis:**} \par        Is the plaintiff entitled to a declaration that he is entitled to one-fourth share in the properties? \dots \par \textbf{**Arguments of Petitioner:**} \par It is contended that the plaintiff can establish his right to cultivate the plots in question, \dots \par \textbf{**Arguments of Respondent:**} \par The respondent's arguments could be that the plaintiff may not be able to establish his right \dots \par
  \textbf{\#\#PREDICTION:} \colorbox{green!20}{The appeal may be dismissed\dots\textbf{(Correct Prediction)}}\par \textbf{\#\#EXPLANATION:}  It can be inferred that the plaintiff's claim to the properties \dots 
  } \\ \hline
  
Phi4-mini-reasoning (16-bit) &
  {
  \textbf{**Legal Issue Analysis:**} \par\ Is the defendant liable? \dots \par  \textbf{**Legal Process Analysis:} \par
  
  \colorbox{orange!10}{%
    \parbox{\dimexpr\linewidth-2\fboxsep}{%
  Respondent Arguments Against Witness Statements \dots art therapy play therapy bibliotherapy music therapy \dots DeFi stablecoins tokenization NFTs metaverse Web3 \dots \textbf{(Random text)} }} } \\ \hline
  Phi4-reasoning (16-bit) &
  { 
  \colorbox{yellow!20}{%
    \parbox{\dimexpr\linewidth-2\fboxsep}{%
    User instructs: "Evaluate whether the appeal should be ACCEPTED (1) or REJECTED (0)"}} \par \colorbox{yellow!20}{%
    \parbox{\dimexpr\linewidth-2\fboxsep}{%
    Now, analyze the case. Let me re-read the case summary carefully.\dots \textbf{(Wrong output format)} }} \par Let\'s check relevant statutes: \dots Also cited precedents: \dots \par 
    \colorbox{purple!15}{%
    \parbox{\dimexpr\linewidth-2\fboxsep}{% 
    Explanation: Because High Court\'s decision \dots But note: The Abolition Act \dots Double-check: Precedent 2 \dots Thus, answer: Prediction: 0 (rejected) \dots \textbf{(Multi-step legal reasoning} }} \par Explanation: Because High Court correctly \dots I\'ll produce answer accordingly.
  \par \textbf{\#\#PREDICTION:} 0 \par \textbf{\#\#EXPLANATION:} The High Court’s \dots
  } \\ \hline
  Qwen3.5-27B (16-bit) &
  {  \textbf{**Legal Issue Analysis:**}  \par What are the grounds for challenging the High Court's judgment \dots \par \textbf{**Arguments of Petitioner:**}  The petitioner may contend that the High Court \dots \par \textbf{**Arguments of Respondent:**}  The respondent's arguments could be that the High Court \dots \par \textbf{\#\#PREDICTIONS:} 
  \colorbox{green!20}{12. The appeal may be dismissed. \dots \textbf{(Correct Prediction)} } \par \textbf{\#\#EXPLANATION:} It can be inferred that a person \dots} \\ \hline
\end{tabularx}%
\caption{Comparative analysis of output generated by different reasoning-oriented models, showing signs of hallucination, repetitive text, sementically redundant text, correct prediction and multi step legal reasoning. }
\label{tab:quantization-effects}
\end{table*}

\end{document}